\documentclass{bmvc2k}

\title{Training-Free Spectral Transductive Refinement for\\ Cross-Domain Few-Shot Classification}

\addauthor{Fahim Rahman$^{*}$}{fahimrahman@iut-dhaka.edu}{1}
\addauthor{S.M. Tanjeeb Meheran Rohan$^{*}$}{tanjeebmeheran@iut-dhaka.edu}{1}
\addauthor{Md. Taimum Ibne Sayed}{taimum@iut-dhaka.edu}{1}
\addauthor{Asaduzzaman Herok}{asaduzzaman34@iut-dhaka.edu}{1}
\addauthor{Md. Bakhtiar Hasan}{bakhtiarhasan@iut-dhaka.edu}{1}

\addinstitution{
 Department of Computer Science and Engineering\\
 Islamic University of Technology\\
 Gazipur, Bangladesh
}

\runninghead{F. Rahman et al.}{Spectral Transductive Refinement}

\usepackage{graphicx}
\usepackage{booktabs}
\usepackage{multirow}
\usepackage{amsmath,amssymb}
\usepackage{algorithm}
\usepackage{algpseudocode}
\usepackage{microtype}
\usepackage{enumitem}
\usepackage{xspace}
\usepackage{array}
\usepackage{tabularx}
\usepackage{float}

\def\etal{\emph{et al}\bmvaOneDot}
\newcommand{\method}{STR\xspace}
\newcommand{\R}{\mathbb{R}}
\newcommand{\Sset}{\mathcal{S}}
\newcommand{\Qset}{\mathcal{Q}}
\newcommand{\papertablefont}{\footnotesize}

\makeatletter
\newcommand{\blfootnote}[1]{%
  \begingroup
  \renewcommand{\thefootnote}{}\footnote{#1}%
  \addtocounter{footnote}{-1}%
  \endgroup
}
\makeatother

\begin{document}
\maketitle

\blfootnote{$^{*}$ Equal contribution.}

\begin{abstract}
Few-shot recognition with frozen visual features is especially fragile under domain shift and one-shot supervision, where a single labelled image is an unreliable estimate of its class. We ask how far this fragility can be reduced \emph{purely at test time}, without retraining the encoder or augmenting the source domain. We present Spectral Transductive Refinement (\method), a training-free transductive inference rule that exploits the geometry of the complete support--query episode. Given frozen embeddings, \method builds a joint $k$-nearest-neighbour graph, maps the episode into a normalized-Laplacian spectral coordinate system, initializes class representatives from the labelled support, and iteratively refines them using pseudo-labelled queries. We evaluate \method under two protocols. A controlled component study with frozen ResNet-18 features shows that spectral refinement consistently improves over single-prototype spectral initialization across five shifted domains, with the largest gains in the one-shot regime where support estimates are weakest. We then benchmark \method against recent Cross-Domain Few-Shot Learning (CD-FSL) methods using the standard \textit{mini}ImageNet-pretrained ResNet-10 backbone over eight established target domains. Operating entirely at inference time, \method attains the highest 1-shot average among compared methods and remains competitive at 5-shot, rivalling approaches relying on heavy source-domain meta-training augmentations. Because \method is transductive, we report its setting explicitly. Diagnostics attribute its gains to iterative refinement in spectral coordinates rather than added prototype capacity, which remains inactive in our configuration.
\end{abstract}

\section{Introduction}
\label{sec:intro}

Few-shot image classification evaluates whether a model can recognize novel classes from only a few labelled examples. The standard episodic protocol, popularized by Matching Networks~\cite{vinyals2016matching} and widely adopted by Prototypical Networks~\cite{snell2017prototypical}, frames evaluation as repeated $N$-way, $K$-shot tasks. In this setting, metric-based methods are appealing because they can classify queries using simple class representatives in a frozen embedding space. However, this simplicity becomes fragile in the one-shot regime: a single support image may be an unreliable representative of its class, especially when the test distribution differs from the data used to train the feature extractor.

A natural way to reduce this brittleness is to use the unlabeled query set available at test time. Transductive few-shot methods exploit this support--query episode structure through graph propagation, label regularization, or information maximization~\cite{liu2019tpn,ziko2020laplacian,boudiaf2020tim,shen2021reranking}. These approaches show that query geometry can improve prediction, but they also make clear that the main challenge is not simply using transduction: the inference rule must extract useful structure without overfitting to noisy episode-level relationships. This issue is particularly important under cross-domain transfer, where frozen features may preserve broad semantics but distort local class geometry.

We study this problem through a training-free inference method called \method, \textbf{S}pectral \textbf{T}ransductive \textbf{R}efinement. Given frozen support and query embeddings, \method constructs a joint $k$-nearest-neighbour graph, computes a normalized-Laplacian spectral representation, initializes class representatives from the labelled support samples, and then iteratively refines these representatives using query pseudo-labels. The spectral representation is intended to provide a compact episode-specific coordinate system in which support and query relationships are smoothed by graph connectivity rather than measured only in the original feature space. \method is transductive by design: it predicts each query batch jointly from the unlabeled target geometry. We treat this as a deliberate modelling choice and report the setting explicitly throughout, so that the comparison against inductive baselines remains transparent.

A key part of our study is to separate the effect of refinement from the effect of additional prototype capacity. The full framework includes an adaptive multi-prototype extension motivated by the possibility that some classes may be multi-modal. Our diagnostics show, however, that under the fixed reported setting this extension typically selects a single representative per class. We are therefore precise about the claim we make: the demonstrated contribution is not that extra prototypes drive the improvement, but that iterative refinement of class representatives in spectral space consistently strengthens the spectral initialization baseline.

We rigorously evaluate \method under two feature protocols. A foundational study using frozen ResNet-18 features isolates the within-framework benefits of transductive refinement under domain shift. To benchmark against the state-of-the-art, we then evaluate \method using a \textit{mini}ImageNet-pretrained ResNet-10 across eight CD-FSL datasets. Across these diverse domains, \method surpasses complex training-time augmentation methods in the 1-shot regime and achieves highly competitive 5-shot performance, all while operating entirely at inference time.

Our contributions are:
\begin{itemize}[leftmargin=*,itemsep=-1pt,topsep=2pt]
    \item We propose \method, a training-free spectral transductive refinement rule that updates class representatives using the joint geometry of support and query samples without fine-tuning the encoder.
    \item We show that \method attains the highest 1-shot average across eight standard CD-FSL benchmarks among compared methods, matching computationally heavy training-time augmentation methods strictly through test-time transductive inference.
    \item We provide a controlled component study showing that spectral refinement improves over spectral initialization most strongly in the one-shot regime, where support-only prototypes are least reliable.
    \item We analyse graph and spectral design choices---including the effect of $k_{\mathrm{nn}}$ on both accuracy and average prototype allocation---and supply mechanism diagnostics that clarify exactly when the more complex parts of the framework activate.
\end{itemize}

\section{Related Work and Positioning}
\label{sec:related}
\subsection{Few-shot recognition and representation quality}
Matching Networks~\cite{vinyals2016matching} introduced an attention-based non-parametric prediction rule and the miniImageNet benchmark that underlies much subsequent evaluation.
Prototypical Networks~\cite{snell2017prototypical} reduced each class to the mean of its support embeddings, providing the starting point for our spectral initialization.
Optimization-based approaches such as MAML~\cite{finn2017maml} adapt model parameters rapidly from support examples, while transfer-style methods emphasize that conventional feature learning and simple classifiers are highly competitive~\cite{wang2019simpleshot,tian2020rethinking,chen2021metabaseline}.
Hu \etal~\cite{hu2022pmf} show that external pretraining and fine-tuning substantially change the apparent difficulty of few-shot benchmarks; Basu \etal~\cite{basu2024strong} further demonstrate strong parameter-efficient transformer adaptation baselines.
These studies heavily inform our evaluation protocol: rather than conflating inference improvements with heterogeneous feature training, we rigorously isolate the impact of our transductive operator by standardizing the frozen encoder (ResNet-18) for our component study, and subsequently matching the exact ResNet-10 source-training protocol of recent competitors for our state-of-the-art comparison.

\subsection{Transductive and graph-based inference}
A transductive classifier may use the complete unlabeled query set when predicting its labels~\cite{chapelle1999transductive}.
In few-shot recognition, Transductive Propagation Networks (TPN) learn an episode graph and propagate support labels to queries~\cite{liu2019tpn}; LaplacianShot introduces a graph Laplacian regularizer for test-time assignment with fixed features~\cite{ziko2020laplacian}; TIM optimizes an information-maximization objective over the query predictions~\cite{boudiaf2020tim}; and re-ranking exploits shared retrieval neighbourhoods~\cite{shen2021reranking}.
Realistic evaluation work additionally warns that transductive assumptions, such as balanced closed-set query batches, should be reported clearly~\cite{veilleux2021realistic}.

Graph transduction also connects to classical semi-supervised learning.
Gaussian fields and local/global consistency propagate labels over a graph under smoothness assumptions~\cite{zhu2003semi,zhou2004learning}; deep semi-supervised label propagation later applied nearest-neighbour graphs constructed from neural representations~\cite{iscen2019labelprop}.
STR differs from direct propagation: it does not solve for graph-smoothed class probabilities. Instead, it uses a spectral embedding of the episode as the coordinate system in which prototype initialization and iterative updates are performed.

\subsection{Spectral geometry and prototype capacity}
Normalized cuts~\cite{shi2000normalized}, spectral clustering~\cite{ng2002spectral}, Laplacian Eigenmaps~\cite{belkin2003laplacian}, and the survey of von Luxburg~\cite{vonluxburg2007tutorial} establish that low-frequency Laplacian eigenvectors encode graph connectivity at a compact scale.
This motivates our use of spectral coordinates when raw frozen features contain local nuisance variation.
Separately, Infinite Mixture Prototypes (IMP) represents a class with an adaptively inferred set of clusters~\cite{allen2019imp}, identifying an important limitation of a single-centroid class model.
We analyse a related episode-adaptive capacity extension, but the measured behaviour requires a narrow interpretation: in the reported fixed setting, adaptive counts do not exceed one representative per class, so the verified contribution is iterative spectral refinement.

\section{Methodology: Spectral Transductive Refinement}
\label{sec:method}
\begin{figure*}[t]
\centering
\includegraphics[width=.98\linewidth]{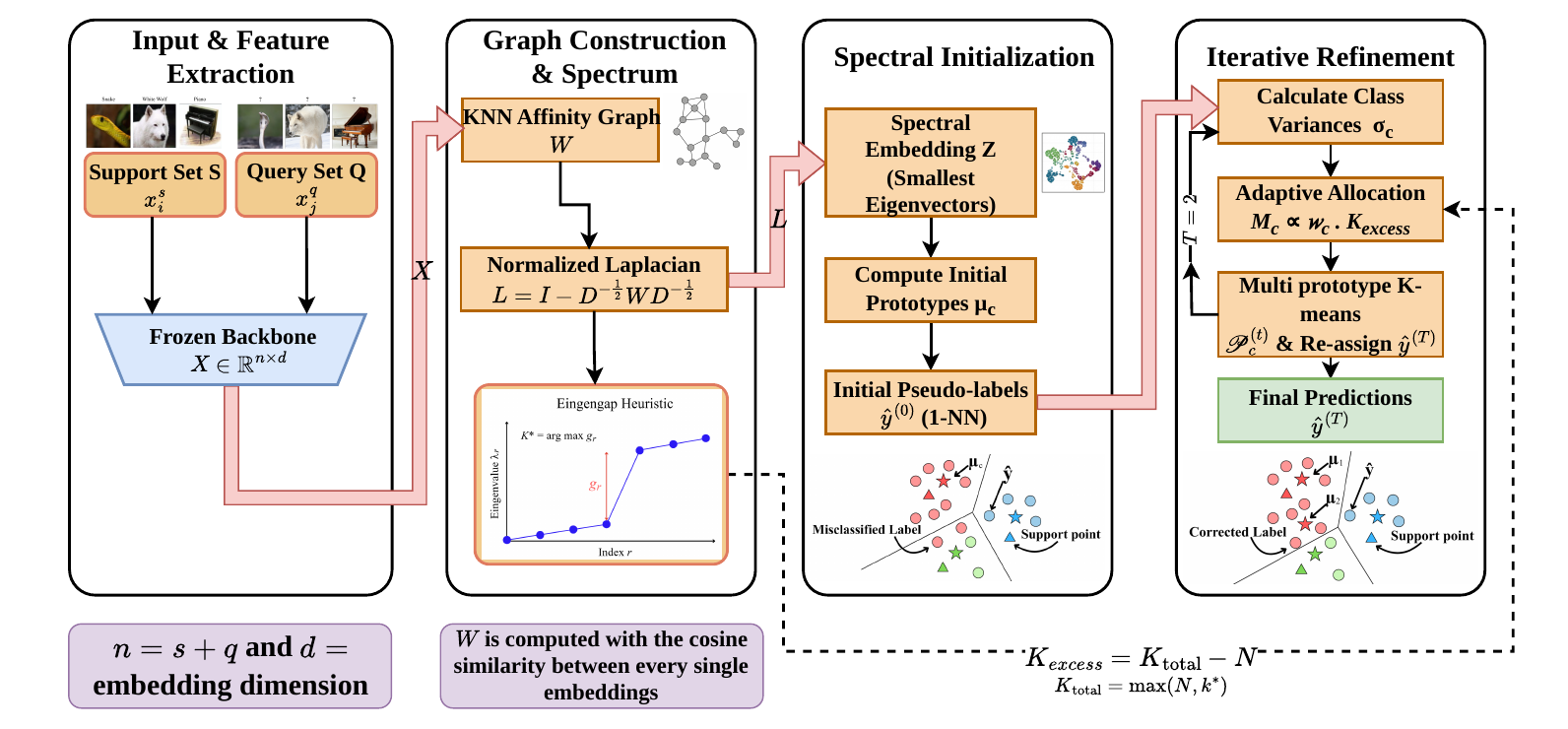}
\caption{\textbf{Overview of the proposed spectral transductive refinement framework and its analysed adaptive-allocation extension.} A frozen visual encoder maps the labelled support set and unlabeled query set to embeddings, from which a joint cosine-similarity $k$NN affinity graph and normalized Laplacian are constructed. The resulting spectral representation is used to initialize class prototypes and query pseudo-labels. These pseudo-labels are then iteratively refined using class-wise spectral structure. The diagram additionally depicts the eigengap-guided, variance-weighted multi-prototype allocation extension; in our experiments, its behaviour is analysed separately so that gains from iterative spectral refinement are not conflated with gains from additional prototype capacity.}
\label{fig:pipeline}
\end{figure*}

We now describe the inference rule carefully, defining all variables before each mathematical step and explaining why that step is required.
An overview is shown in Fig.~\ref{fig:pipeline}.

\subsection{Task definition and notation}
\label{sec:task}
An \emph{episode} is a small classification task with $N$ novel classes.
For each class, $K$ labelled images are provided; hence $K$ is the number of \emph{shots} and $N$ is the number of \emph{ways}.
Let $n_s=NK$ denote the number of labelled support images and let $n_q$ denote the number of unlabeled query images whose labels must be predicted.
We use $\mathbf{x}_i^s$ for the $i$-th support image, $y_i^s \in \{1,\ldots,N\}$ for its class label, and $\mathbf{x}_j^q$ for the $j$-th query image.
The support and query sets are
\begin{equation}
\Sset = \{(\mathbf{x}_i^s,y_i^s)\}_{i=1}^{n_s},
\qquad
\Qset = \{\mathbf{x}_j^q\}_{j=1}^{n_q},
\qquad n_s = NK.
\label{eq:episode}
\end{equation}
In inductive inference each query is processed independently.
STR instead follows the transductive setting: all $n_q$ query images in an episode are visible jointly at inference time, but their labels are never used.
This is necessary because the method aims to infer episode-level geometry from the unlabeled query distribution.

Let $\phi(\cdot)$ denote a frozen visual encoder and let $d$ be its output feature dimension.
The support and query embedding matrices are
\begin{equation}
\mathbf{S}= [\phi(\mathbf{x}_1^s),\ldots,\phi(\mathbf{x}_{n_s}^s)]^{\top}
\in\R^{n_s\times d},
\qquad
\mathbf{Q}= [\phi(\mathbf{x}_1^q),\ldots,\phi(\mathbf{x}_{n_q}^q)]^{\top}
\in\R^{n_q\times d}.
\label{eq:separate_features}
\end{equation}
We concatenate these matrices as
\begin{equation}
\mathbf{X}=\begin{bmatrix}\mathbf{S}\\\mathbf{Q}\end{bmatrix}
\in\R^{n\times d},
\qquad n=n_s+n_q.
\label{eq:feature_matrix}
\end{equation}
No gradient is computed and no parameter of $\phi$ is updated.
Concatenation is required because the graph in the next step must encode support--support, support--query, and query--query relationships within the same episode.

\begin{table}[t]
\centering
\papertablefont
\caption{Notation used in Sec.~\ref{sec:method}.}
\label{tab:notation}
\setlength{\tabcolsep}{3.4pt}
\begin{tabular}{lp{6.4cm}}
\toprule
Symbol & Meaning \\
\midrule
$N$, $K$ & Number of classes and labelled support images per class. \\
$n_s{=}NK$, $n_q$ & Numbers of support and query images in one episode. \\
$\phi$, $d$ & Frozen encoder and its feature dimension. \\
$\mathbf{X}$, $n$ & Joint feature matrix and total episode size $n_s+n_q$. \\
$k_{\mathrm{nn}}$ & Number of graph neighbours retained per image. \\
$\mathbf{W}$, $\mathbf{D}$, $\mathbf{L}$ & Affinity, degree, and normalized Laplacian matrices. \\
$d_{\mathrm{spec}}$, $\mathbf{Z}$ & Spectral dimension and spectral episode coordinates. \\
$\boldsymbol{\mu}_c^{(t)}$ & Representative of class $c$ at refinement iteration $t$. \\
$\hat{y}_j^{(t)}$, $T$ & Query pseudo-label after iteration $t$ and total refinements. \\
\bottomrule
\end{tabular}
\end{table}

\subsection{Joint neighbourhood graph}
\label{sec:graph}
Let $\mathbf{x}_i \in \R^d$ and $\mathbf{x}_j \in \R^d$ denote two rows of the joint feature matrix $\mathbf{X}$.
We use cosine similarity because frozen representations may differ in feature magnitude while direction captures semantic proximity.
The affinity between two embeddings is
\begin{equation}
 s_{ij}=\frac{\mathbf{x}_i^{\top}\mathbf{x}_j}
 {\|\mathbf{x}_i\|_2\|\mathbf{x}_j\|_2+\epsilon},
\label{eq:cosine}
\end{equation}
where $\epsilon>0$ is a numerical-stability constant.
For each node $i$, we retain its $k_{\mathrm{nn}}$ most similar neighbours and symmetrize the resulting sparse affinities to obtain $\mathbf{W}\in\R^{n\times n}$.
Here $k_{\mathrm{nn}}$ controls locality: a graph that is too sparse may disconnect relevant query structure, whereas a graph that is too dense can introduce cross-class connections.
This motivates the neighbourhood ablation in Sec.~\ref{sec:analysis}.

Let $\mathbf{D}\in\R^{n\times n}$ be the diagonal degree matrix with $D_{ii}=\sum_j W_{ij}$, and let $\mathbf{I}$ be the $n\times n$ identity matrix.
We construct the symmetric normalized graph Laplacian
\begin{equation}
\mathbf{L}=\mathbf{I}-\mathbf{D}^{-1/2}\mathbf{W}\mathbf{D}^{-1/2}.
\label{eq:laplacian}
\end{equation}
The normalized form is necessary because episode graphs can contain non-uniform local density; degree normalization prevents highly connected points from dominating the representation merely because they have larger neighbourhood mass~\cite{shi2000normalized,vonluxburg2007tutorial}.

\subsection{Spectral episode coordinates}
\label{sec:spectral}
Let $(\lambda_r,\mathbf{v}_r)$ denote the $r$-th eigenpair of $\mathbf{L}$, sorted so that $\lambda_1\le \lambda_2\le\cdots\le\lambda_n$.
For a connected graph, the first eigenvector $\mathbf{v}_1$ is the trivial constant mode and does not distinguish samples.
Let $d_{\mathrm{spec}}$ denote the selected spectral dimension.
STR represents each episode sample using the next $d_{\mathrm{spec}}$ low-frequency eigenvectors:
\begin{equation}
\mathbf{Z}=[\mathbf{v}_2,\mathbf{v}_3,\ldots,\mathbf{v}_{d_{\mathrm{spec}}+1}]
\in\R^{n\times d_{\mathrm{spec}}}.
\label{eq:spectral}
\end{equation}
The first $n_s$ rows of $\mathbf{Z}$ are denoted $\mathbf{Z}_s$ and the remaining $n_q$ rows are denoted $\mathbf{Z}_q$.
A row $\mathbf{z}_i^s$ is therefore the spectral representation of a labelled support image, while $\mathbf{z}_j^q$ is the spectral representation of an unlabeled query.
This step is necessary because raw encoder features express individual visual similarity, while low-frequency spectral coordinates emphasize connectivity in the joint episode graph: queries repeatedly connected through local neighbourhoods become easier to summarize by shared representatives.
We treat this as an empirical design hypothesis, tested by component comparisons, rather than claiming that spectral coordinates are universally optimal.

\subsection{Spectral initialization}
\label{sec:init}
For each class index $c\in\{1,\ldots,N\}$, let $\mathcal{I}_c=\{i\,|\,y_i^s=c\}$ be the indices of labelled support images belonging to class $c$.
The initial representative of class $c$ is the mean of its support points in spectral coordinates:
\begin{equation}
\boldsymbol{\mu}_c^{(0)}
=\frac{1}{|\mathcal{I}_c|}\sum_{i\in\mathcal{I}_c}\mathbf{z}_i^s.
\label{eq:init_proto}
\end{equation}
Each query receives an initial pseudo-label through its closest spectral representative:
\begin{equation}
\hat{y}_j^{(0)}
=\arg\min_{c\in\{1,\ldots,N\}}
\|\mathbf{z}_j^q-\boldsymbol{\mu}_c^{(0)}\|_2^2.
\label{eq:init_assign}
\end{equation}
This initialization is intentionally conservative: it uses only true support labels to create the first class summaries.
It also yields the \emph{spectral single-prototype} baseline used to measure whether subsequent query-driven refinement adds value.

\subsection{Iterative representative refinement}
\label{sec:refinement}
Let $T$ denote the number of refinement iterations, and let $t\in\{1,\ldots,T\}$ denote the current iteration.
Given predictions from the preceding step, STR forms a class-conditioned set containing both labelled support points and query points currently assigned to class $c$:
\begin{equation}
\mathcal{X}_c^{(t)}=
\{\mathbf{z}_i^s\mid y_i^s=c\}
\cup
\{\mathbf{z}_j^q\mid\hat{y}_j^{(t-1)}=c\}.
\label{eq:class_set}
\end{equation}
The updated representative is
\begin{equation}
\boldsymbol{\mu}_c^{(t)}
=\frac{1}{|\mathcal{X}_c^{(t)}|}
\sum_{\mathbf{z}\in\mathcal{X}_c^{(t)}}\mathbf{z},
\label{eq:refined_proto}
\end{equation}
and the queries are reassigned as
\begin{equation}
\hat{y}_j^{(t)}
=\arg\min_{c\in\{1,\ldots,N\}}
\|\mathbf{z}_j^q-\boldsymbol{\mu}_c^{(t)}\|_2^2.
\label{eq:refine_assign}
\end{equation}
The refinement step is necessary in the one-shot regime because Eq.~\eqref{eq:init_proto} begins from only one labelled point per class.
When the spectral initialization is reasonably accurate, pseudo-labelled queries provide additional evidence for the episode-specific location of each class.
Conversely, Eq.~\eqref{eq:class_set} also makes the failure mode explicit: inaccurate initial pseudo-labels can pull a representative toward an incorrect cluster.

\begin{algorithm}[t]
\caption{Spectral Transductive Refinement (STR)}
\label{alg:str}
\begin{algorithmic}[1]
\Require Support embeddings $\mathbf{S}$ and labels $\mathbf{y}^s$; query embeddings $\mathbf{Q}$; neighbourhood size $k_{\mathrm{nn}}$; spectral dimension $d_{\mathrm{spec}}$; iterations $T$
\Ensure Query predictions $\hat{\mathbf{y}}^{(T)}$
\State Concatenate $\mathbf{X}=[\mathbf{S};\mathbf{Q}]$
\State Construct $k$NN graph $\mathbf{W}$ and normalized Laplacian $\mathbf{L}$
\State Compute $\mathbf{Z}$ from the non-trivial low-frequency eigenvectors of $\mathbf{L}$
\State Initialize $\hat{\mathbf{y}}^{(0)}$ using Eqs.~\eqref{eq:init_proto}--\eqref{eq:init_assign}
\For{$t=1,\ldots,T$}
    \For{$c=1,\ldots,N$}
        \State Form $\mathcal{X}_c^{(t)}$ using Eq.~\eqref{eq:class_set}
        \State Update $\boldsymbol{\mu}_c^{(t)}$ using Eq.~\eqref{eq:refined_proto}
    \EndFor
    \State Reassign all queries using Eq.~\eqref{eq:refine_assign}
\EndFor
\State \Return $\hat{\mathbf{y}}^{(T)}$
\end{algorithmic}
\end{algorithm}

\subsection{Analysed multi-prototype extension}
\label{sec:extension}
The full pipeline diagram includes an exploratory extension intended to increase class capacity when an episode exhibits multi-modal spectral structure.
Let $\lambda_r$ denote the ordered eigenvalues already defined in Sec.~\ref{sec:spectral}.
For consecutive non-trivial modes, define an eigengap $g_r=\lambda_{r+2}-\lambda_{r+1}$ and let $k^\star=\arg\max_r g_r$ denote the index of the largest candidate gap.
The extension uses
\begin{equation}
P_{\mathrm{ref}}=\max(N,k^\star),
\qquad
P_{\mathrm{extra}}=P_{\mathrm{ref}}-N
\label{eq:budget}
\end{equation}
as an episode-level reference for extra prototype capacity.
To distribute that capacity, let $\overline{\mathbf{z}}_c$ be the mean of $\mathcal{X}_c^{(t)}$ and let
\begin{equation}
\sigma_c^2=\frac{1}{|\mathcal{X}_c^{(t)}|}
\sum_{\mathbf{z}\in\mathcal{X}_c^{(t)}}
\|\mathbf{z}-\overline{\mathbf{z}}_c\|_2^2,
\qquad
w_c=\frac{\sigma_c^2}{\sum_{c'=1}^{N}\sigma_{c'}^2}.
\label{eq:variance}
\end{equation}
Here $\sigma_c^2$ measures current spectral dispersion and $w_c$ is its normalized share across classes.
The recorded implementation assigns approximately variance-proportional counts $M_c\geq1$ and clusters $\mathcal{X}_c^{(t)}$ into $M_c$ representatives before reassignment.
This step was motivated by prior multi-prototype modelling~\cite{allen2019imp}: a visually broad class may be poorly summarized by a single centroid.
We nonetheless judge it strictly by its recorded behaviour.
In the fixed reported setting, $M_c=1$ for every class in the structured experiments (Table~\ref{tab:mechanism_diag}); hence the extension reduces exactly to \method and is not claimed as the source of improvement. Section~\ref{sec:knn_proto_activation} further examines how the allocation changes when graph locality is varied.

\subsection{Computational cost}
For $n$ episode samples of dimension $d$, pairwise cosine similarities cost $\mathcal{O}(n^2d)$. Exact eigendecomposition costs $\mathcal{O}(n^3)$, while representative updates require only $\mathcal{O}(Tnd_{\mathrm{spec}})$ work in the one-prototype STR regime. Because standard 5-way episodes are small, spectral decomposition is highly feasible, though it adds overhead relative to direct prototype matching.

\section{Experimental Protocol}
\label{sec:protocol}
\subsection{Episode construction and reporting}
The standard miniImageNet episodic evaluation originated with Matching Networks~\cite{vinyals2016matching} and was adopted by Prototypical Networks~\cite{snell2017prototypical}.
Following this established protocol, we evaluate 5-way tasks with $K\in\{1,5\}$ labelled support images per class and 15 unlabeled query images per class.
Each reported value is the mean classification accuracy over 600 test episodes with a 95\% confidence interval.
For context, tieredImageNet was introduced with semantically separated splits by Ren \etal~\cite{ren2018semisupervised}; the revised central analysis uses miniImageNet only as an in-domain anchor and focuses on target-domain component generalization.

\subsection{Frozen representation and targets}
\label{sec:targets}
To thoroughly evaluate component generalization, our study utilizes two distinct feature extraction protocols. First, for our foundational component study, we use a standard pretrained frozen ResNet-18~\cite{he2016deep}, evaluated on five shifted domains: Caltech-101~\cite{feifei2004caltech}, DTD~\cite{cimpoi2014dtd}, FC100~\cite{oreshkin2018tadam}, Omniglot~\cite{lake2015human}, and Oxford-102 Flowers~\cite{nilsback2008flowers}.

Second, to ensure a rigorous comparison against established Cross-Domain Few-Shot Learning (CD-FSL) state-of-the-art methods, we employ a standard ResNet-10~\cite{he2016deep} backbone pretrained exclusively on the \textit{mini}ImageNet source domain. For this protocol, we evaluate on eight standard benchmarks: the BSCD-FSL collection (ChestX~\cite{wang2017chestx}, ISIC~\cite{tschandl2018ham10000}, EuroSAT~\cite{helber2019eurosat}, CropDisease~\cite{mohanty2016using}) and the \textit{mini}-CUB collection (CUB~\cite{wah2011cub}, Cars~\cite{krause20133d}, Places~\cite{zhou2017places}, Plantae~\cite{van2018inaturalist}).

\subsection{Configuration and comparison scope}
\label{sec:configuration}
The structured component study uses $k_{\mathrm{nn}}=20$, $d_{\mathrm{spec}}=5$, and $T=2$ refinement iterations. The ResNet-18 experiments isolate the within-framework effect of our inference rule across shifted domains. Subsequently, the ResNet-10 experiments allow us to directly contextualize our results against recent state-of-the-art training-time augmentation methods, specifically Meta-Exploiting Frequency Prior~\cite{zhou2024meta} and Harmonized Amplitude Perturbation (HAP)~\cite{li2025hap}. We emphasise at the outset that \method is transductive whereas these baselines follow the standard inductive CD-FSL protocol; we state this difference openly and return to it when interpreting the results (Sec.~\ref{sec:cross_domain}, Sec.~\ref{sec:comparison_scope}).

\section{Results and Analysis}
\label{sec:results}
\subsection{Published standard-benchmark context}
\label{sec:published_context}
\begin{table*}[t]
\centering
\papertablefont
\setlength{\tabcolsep}{2pt}
\renewcommand{\arraystretch}{1.07}
\caption{Published standard-benchmark context on miniImageNet and tieredImageNet. Unless marked as ours, values are reported by the cited publications. Methods are separated by their stated feature-training protocol; our frozen ResNet-18 uses supervised ImageNet-1K pretraining, so the table is contextual rather than a fully controlled same-feature comparison.}
\label{tab:published_standard}
\begin{tabular*}{\textwidth}{@{\extracolsep{\fill}}llcccc@{}}
\toprule
Method & Reported feature protocol &
\multicolumn{2}{c}{miniImageNet} &
\multicolumn{2}{c}{tieredImageNet} \\
\cmidrule(lr){3-4}\cmidrule(lr){5-6}
& & 1-shot & 5-shot & 1-shot & 5-shot \\
\midrule
SimpleShot~\cite{wang2019simpleshot} & ResNet-18 & 62.9 & 80.0 & 68.9 & 84.6 \\
LaplacianShot~\cite{ziko2020laplacian} & ResNet-18 & 72.1 & 82.3 & 79.0 & 86.4 \\
TIM-ADM~\cite{boudiaf2020tim} & ResNet-18 & 73.6 & 85.0 & 80.0 & 88.5 \\
TIM-GD~\cite{boudiaf2020tim} & ResNet-18 & 73.9 & 85.0 & 79.9 & 88.5 \\
\midrule
Spectral init. (ours) & ResNet-18 & 85.62$\pm$0.64 & 92.86$\pm$0.32 & 87.21$\pm$0.64 & 95.40$\pm$0.34 \\
\textbf{STR (ours)} & ResNet-18 & \textbf{87.55$\pm$0.69} & \textbf{93.21$\pm$0.33} & \textbf{92.58$\pm$0.62} & \textbf{96.31$\pm$0.34} \\
Matched gain & -- & +1.93 & +0.35 & +5.37 & +0.91 \\
\bottomrule
\multicolumn{6}{l}{\papertablefont Published rows follow their cited reporting protocols; our rows use 600 episodes.}
\end{tabular*}
\end{table*}

Table~\ref{tab:published_standard} reports standard-benchmark context grouped by the stated feature protocol. Values for existing methods are taken from the cited publications, while the spectral-initialization and STR rows are evaluated within our pipeline. The reported STR improvements over its matched spectral initialization are $+1.93$ and $+0.35$ points on miniImageNet and $+5.37$ and $+0.91$ points on tieredImageNet in the 1-shot and 5-shot settings, respectively. Two caveats keep this table strictly contextual. First, the representation-training protocols differ across blocks: our frozen ResNet-18 relies on standard pretraining and is therefore a substantially stronger feature extractor than the base-class-trained backbones behind the published rows, which explains the higher absolute accuracies. Second, miniImageNet classes are themselves drawn from ImageNet, so these in-domain rows benefit from source overlap and should not be read as a controlled ranking against published numbers. We therefore use this table only as internal context and base our cross-domain claims on the shifted-domain study (Sec.~\ref{sec:cross_domain}) and the CD-FSL benchmarks (Table~\ref{tab:crossdomain_8datasets}), where the encoder never observes the target domains.

\subsection{Component generalization under domain shift}
\label{sec:cross_domain}

\paragraph{Foundation-feature transfer (ResNet-18).}
\begin{table*}[t]
\centering
\papertablefont
\setlength{\tabcolsep}{4.3pt}
\renewcommand{\arraystretch}{1.07}
\caption{Cross-domain component generalization using the same standard pretrained frozen ResNet-18 feature extractor. Spectral initialization is compared with STR under the matched episodic protocol. Results are mean accuracy (\%) with 95\% confidence intervals over 600 episodes.}
\label{tab:crossdomain_components}
\begin{tabular*}{\textwidth}{@{\extracolsep{\fill}}lcccccc@{}}
\toprule
\multirow{2}{*}{Dataset} & \multicolumn{3}{c}{5-way 1-shot} & \multicolumn{3}{c}{5-way 5-shot}\\
\cmidrule(lr){2-4}\cmidrule(lr){5-7}
& Spectral init. & STR & $\Delta$ & Spectral init. & STR & $\Delta$\\
\midrule
Caltech101 & 89.57$\pm$0.55 & 94.97$\pm$0.48 & +5.40 & 97.30$\pm$0.19 & 98.09$\pm$0.17 & +0.79 \\
DTD & 57.28$\pm$0.95 & 60.27$\pm$1.11 & +2.99 & 73.39$\pm$0.71 & 73.78$\pm$0.76 & +0.39 \\
Oxford-102 & 83.98$\pm$0.61 & 90.49$\pm$0.65 & +6.51 & 94.66$\pm$0.33 & 95.78$\pm$0.31 & +1.12 \\
FC100 & 55.27$\pm$0.99 & 58.08$\pm$1.10 & +2.81 & 67.86$\pm$0.85 & 68.30$\pm$0.88 & +0.44 \\
Omniglot & 77.80$\pm$0.86 & 83.31$\pm$0.93 & +5.51 & 90.95$\pm$0.50 & 91.72$\pm$0.53 & +0.77 \\
\midrule
Mean gain & -- & -- & \textbf{+4.64} & -- & -- & \textbf{+0.70} \\
\bottomrule
\end{tabular*}
\end{table*}

Table~\ref{tab:crossdomain_components} reveals a pronounced one-shot pattern. Across Caltech-101, DTD, FC100, Omniglot, and Oxford-102, STR improves over spectral initialization by $+2.81$ to $+6.51$ percentage points, yielding a mean gain of $+4.64$ points. The effect is smaller in 5-shot evaluation, where five labelled supports already provide a more stable class representative: the mean cross-domain gain is $+0.70$ points. These results support the interpretation that query-guided refinement provides the maximum geometric benefit when the initial labelled estimate is weak.

The strongest per-dataset improvements occur on Oxford-102 and Omniglot. These domains exhibit substantial appearance variability in fine-grained flowers and character shapes, respectively, so a representative initialized from a single support image can be particularly brittle. By contrast, the smaller gains on DTD and FC100 indicate that refinement is not uniformly sufficient to overcome limited class separability in the frozen representation.

\begin{figure*}[t]
\centering
\includegraphics[width=0.68\textwidth]{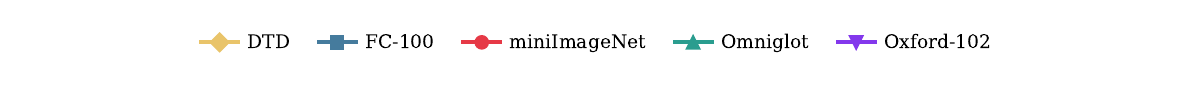}\\[-1mm]
\vspace{1mm}

\begin{minipage}[t]{0.32\textwidth}\centering
  \includegraphics[width=\linewidth]{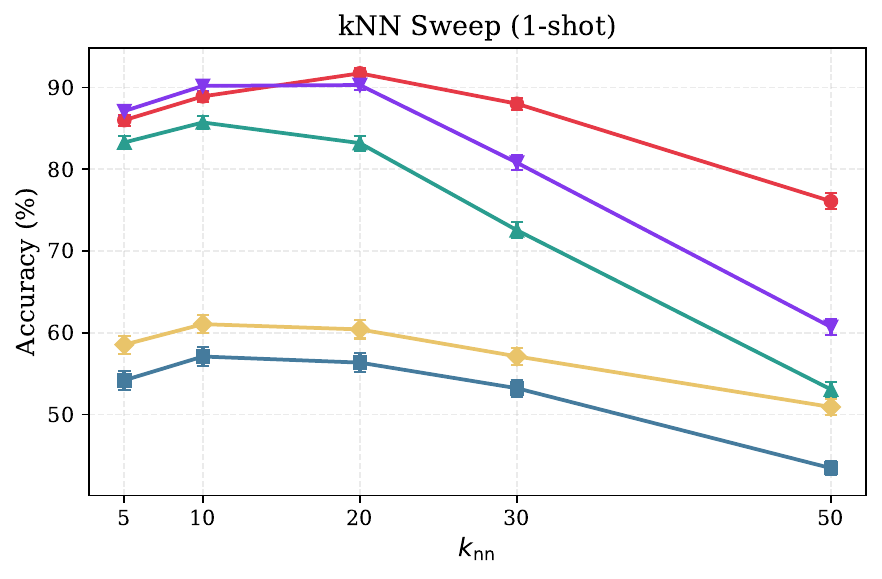}\\[-1mm]
  {\footnotesize (a) $k_{\mathrm{nn}}$ sweep.}
\end{minipage}\hfill
\begin{minipage}[t]{0.32\textwidth}\centering
  \includegraphics[width=\linewidth]{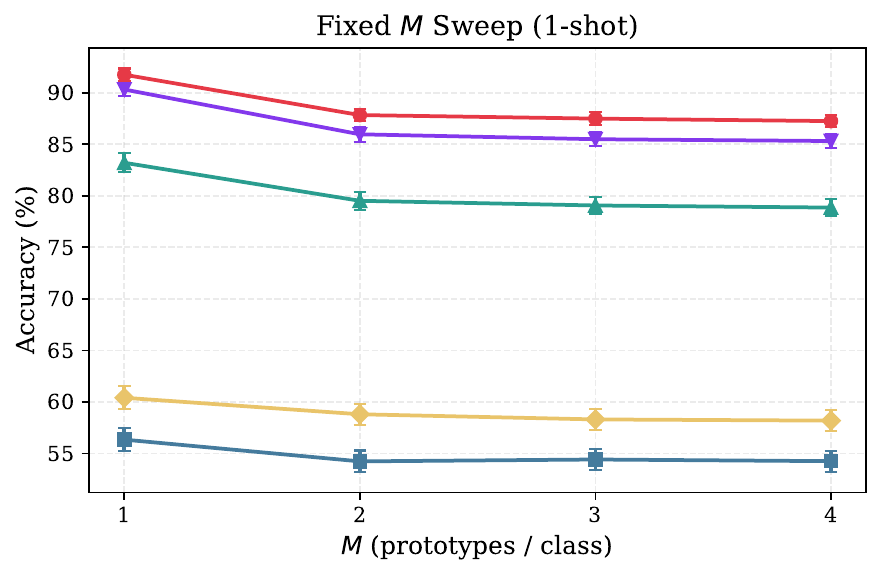}\\[-1mm]
  {\footnotesize (b) Fixed prototype count.}
\end{minipage}\hfill
\begin{minipage}[t]{0.32\textwidth}\centering
  \includegraphics[width=\linewidth]{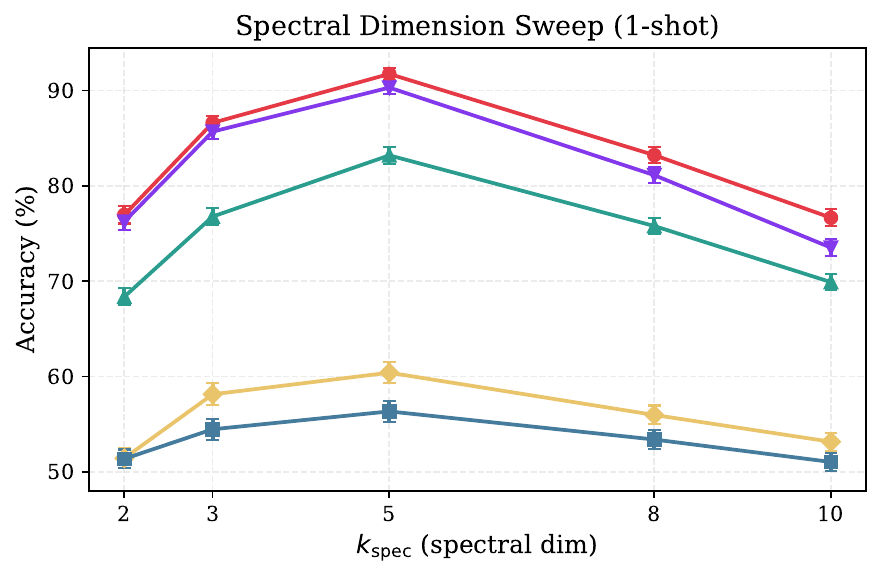}\\[-1mm]
  {\footnotesize (c) Spectral dimension.}
\end{minipage}

\caption{One-shot sensitivity analysis with frozen ResNet-18 features over the component-study datasets. The shared legend above the plots identifies the datasets used in all three sweeps. The $k_{\mathrm{nn}}$ sweep tests the locality of the joint graph, the fixed-$M$ sweep tests whether additional centroids improve refinement, and the $d_{\mathrm{spec}}$ sweep tests the dimensionality of the spectral coordinate system.}
\label{fig:ablation_k1}
\end{figure*}

\paragraph{Standard CD-FSL benchmarks (ResNet-10).}
To further validate our method against current state-of-the-art CD-FSL architectures, we evaluate \method using the standard \textit{mini}ImageNet-pretrained ResNet-10 backbone across eight established domains.

\begin{table*}[h!]
\centering
\papertablefont
\caption{\textbf{Cross-domain few-shot accuracy (\%) across eight standard benchmarks}, using a ResNet-10 backbone pretrained on the \textit{mini}ImageNet source domain. \method establishes strong cross-domain performance strictly through test-time inference, attaining the highest 1-shot average among the compared methods and remaining competitive at 5-shot. \textbf{Setting note:} the cited baselines are inductive training-time methods, whereas \method is transductive (it predicts each query batch jointly from the unlabeled target episode); we report this distinction openly and discuss it in Sec.~\ref{sec:cross_domain} and Sec.~\ref{sec:comparison_scope}. The ``Modification Phase'' column indicates whether adaptation occurs during source-domain training or only at inference.}
\label{tab:crossdomain_8datasets}
\resizebox{\textwidth}{!}{
\begin{tabular}{l c | cccc | cccc | c}
\toprule
\multirow{2}{*}{\textbf{Method}} & \textbf{Modification} & \multicolumn{4}{c|}{\textbf{mini-CUB Domains}} & \multicolumn{4}{c|}{\textbf{BSCD-FSL Domains}} & \multirow{2}{*}{\textbf{Avg.}} \\
\cmidrule(lr){3-6} \cmidrule(lr){7-10}
 & \textbf{Phase} & \textbf{CUB} & \textbf{Cars} & \textbf{Places} & \textbf{Plantae} & \textbf{ChestX} & \textbf{ISIC} & \textbf{EuroSAT} & \textbf{CropDisease} & \\
\midrule
\multicolumn{11}{c}{\textbf{5-way 1-shot}} \\
\midrule
GNN~\cite{garcia2018few} & Training & 45.69 & 31.79 & 53.10 & 35.60 & 22.00 & 32.02 & 63.69 & 64.48 & 43.55 \\
ATA~\cite{wang2021cross} & Training & 45.00 & 33.61 & 53.57 & 34.42 & 22.10 & 33.21 & 61.35 & 67.47 & 43.84 \\
DSU~\cite{li2022uncertainty} & Training & 47.74 & 37.19 & 54.81 & 31.61 & 22.35 & 31.43 & 64.55 & 64.73 & 44.30 \\
FLOR~\cite{zou2024flatten} & Training & 49.99 & 37.41 & 49.75 & 40.12 & 23.11 & \textbf{38.11} & 62.95 & 71.34 & 46.60 \\
StyleAdv~\cite{fu2023styleadv} & Training & 48.49 & 34.64 & 58.58 & 41.13 & 22.64 & 33.96 & \textbf{70.94} & 74.13 & 48.06 \\
FreqPrior~\cite{zhou2024meta} & Training & 51.55 & 37.04 & 52.06 & 41.55 & 22.82 & 33.98 & 64.31 & 71.47 & 46.85 \\
HAP~\cite{li2025hap} & Training & 45.73 & 33.18 & 54.09 & 38.15 & 23.01 & 35.68 & 66.56 & 67.05 & 45.43 \\
\textbf{\method (Ours)} & \textbf{Inference} & \textbf{56.47} & \textbf{43.54} & \textbf{67.34} & \textbf{43.19} & \textbf{24.75} & 28.94 & 68.98 & \textbf{79.59} & \textbf{51.60} \\

\midrule
\multicolumn{11}{c}{\textbf{5-way 5-shot}} \\
\midrule
GNN~\cite{garcia2018few} & Training & 62.25 & 44.28 & 70.84 & 52.53 & 25.27 & 43.94 & 83.64 & 87.96 & 58.84 \\
DSU~\cite{li2022uncertainty} & Training & 67.94 & 45.65 & 75.17 & 54.31 & 24.70 & 45.85 & 83.08 & 86.30 & 59.99 \\
ATA~\cite{wang2021cross} & Training & 66.22 & 49.14 & 75.48 & 52.69 & 24.32 & 44.91 & 83.75 & 90.59 & 60.89 \\
FLOR~\cite{zou2024flatten} & Training & 70.39 & 53.43 & 68.51 & 55.81 & 26.71 & \textbf{51.44} & 80.87 & 89.02 & 62.02 \\
StyleAdv~\cite{fu2023styleadv} & Training & 68.72 & 50.13 & 77.73 & \textbf{61.52} & 26.07 & 45.77 & \textbf{86.58} & \textbf{93.65} & \textbf{63.77} \\
FreqPrior~\cite{zhou2024meta} & Training & \textbf{73.61} & 54.22 & 73.78 & 61.39 & 26.53 & 48.70 & 81.24 & 90.68 & \textbf{63.77} \\
HAP~\cite{li2025hap} & Training & 62.49 & 47.61 & 73.07 & 56.54 & 26.32 & 48.26 & 84.88 & 89.49 & 61.08 \\
\textbf{\method (Ours)} & \textbf{Inference} & 71.16 & \textbf{57.70} & \textbf{82.42} & 58.52 & \textbf{28.52} & 35.30 & 81.66 & 91.42 & 63.34 \\
\bottomrule
\end{tabular}
}
\end{table*}

As shown in Table~\ref{tab:crossdomain_8datasets}, \method attains the highest 1-shot average across the eight domains ($51.60\%$), exceeding recent training-time methods such as FreqPrior~\cite{zhou2024meta} and HAP~\cite{li2025hap}, and remains highly competitive at 5-shot ($63.34\%$), trailing only FreqPrior and StyleAdv. Two properties of this comparison deserve explicit emphasis. First, the competing methods address domain shift \emph{during source-domain meta-training} through complex frequency-domain augmentations, adversarial style perturbations, or feature-space regularization, whereas \method introduces no training-time machinery and modifies no encoder weight: its entire adaptation occurs at inference. Second, \method is \emph{transductive}---it predicts each query batch jointly using the unlabeled target geometry---while the cited baselines follow the standard inductive CD-FSL protocol. We state this distinction openly rather than obscure it. Our claim is therefore not that transduction and induction are interchangeable, but that exploiting the unlabeled target episode at test time is a remarkably cheap and effective route to cross-domain robustness, most decisively when labelled data is scarcest (1-shot).

The single clear exception is ISIC, where \method underperforms the strongest baselines in both regimes. This boundary case is informative rather than incidental: dermoscopic skin-lesion images lie far from the ImageNet source manifold, so the frozen features yield a poor initial spectral partition, and---consistent with the failure mode made explicit in Eq.~\eqref{eq:class_set} and Sec.~\ref{sec:limitations}---iterative pseudo-labelling then amplifies rather than corrects these errors. The pattern reinforces our interpretation of \emph{when} the method helps: refinement is most valuable when the frozen representation already places classes in broadly separable positions that the spectral update can sharpen, and least valuable when the initial partition is already unreliable.

\subsection{What mechanism produces the gain?}
\begin{table}[t]
\centering
\papertablefont
\setlength{\tabcolsep}{3.0pt}
\renewcommand{\arraystretch}{1.07}
\caption{One-shot mechanism diagnostic under the fixed configuration
($k_{\mathrm{nn}}=20$, $d_{\mathrm{spec}}=5$, $T=2$). The allocation
extension uses one prototype per class, while forcing $M=2$ is less reliable.}
\label{tab:mechanism_diag}
\begin{tabular}{lrrrr}
\toprule
Dataset & Init. & Fixed $M{=}2$ & STR & $\overline{M}_c$\\
\midrule
Caltech101 & 89.57 & 92.40 & 94.97 & 1.000 \\
DTD & 57.28 & 58.60 & 60.27 & 1.000 \\
Oxford-102 & 83.98 & 87.10 & 90.49 & 1.000 \\
FC100 & 55.27 & 54.30 & 58.08 & 1.000 \\
Omniglot & 77.80 & 79.50 & 83.31 & 1.000 \\
\bottomrule
\end{tabular}
\end{table}

Table~\ref{tab:mechanism_diag} addresses the central interpretability question.
At $k_{\mathrm{nn}}=20$, $d_{\mathrm{spec}}=5$, and $T=2$, the logged extension reports an average of exactly one representative per class on all displayed datasets.
Its performance consequently coincides with iterative single-prototype refinement.
Moreover, forcing $M=2$ produces lower accuracy than STR on every displayed one-shot dataset.
The evidence therefore does not support attributing performance to added centroids; it supports the iterative update of one class representative in spectral coordinates.

This diagnostic also connects back to IMP~\cite{allen2019imp}.
Adaptive prototype capacity remains a reasonable direction when class structure is truly multi-modal, but it is not sufficient that a pipeline contains an allocation rule: the allocation must activate and improve predictions in the evaluated regime.
Our analysis makes this distinction explicit, and the method name intentionally reflects only the demonstrated mechanism.

\subsection{Sensitivity to graph and spectral construction}
\label{sec:analysis}

Figure~\ref{fig:ablation_k1}(a) shows that graph construction is not innocuous.
Small neighbourhoods can omit within-class connectivity, whereas large neighbourhoods increasingly connect visually similar but label-incompatible samples.
A moderate graph size is consequently used in the fixed component study.
Figure~\ref{fig:ablation_k1}(c) exhibits a similar trade-off in spectral dimension: too few eigenvectors underspecify episode connectivity, while high-dimensional spectral coordinates retain less stable structure.

The fixed-$M$ sweep in Fig.~\ref{fig:ablation_k1}(b) is especially important for the revised interpretation.
On these data, additional representatives do not translate into better predictions: premature fragmentation of a pseudo-labelled set can create noisy class centres rather than capture meaningful modes.
Thus, the fixed-$M$ ablation agrees with the prototype-allocation diagnostics and motivates focusing the empirical contribution on spectral transductive refinement rather than on additional prototype capacity.

\subsection{Graph locality and prototype activation}
\label{sec:knn_proto_activation}

\begin{figure}[t]
\centering
\includegraphics[width=0.68\textwidth]{figures/ablations/shared_legend.pdf}\\[-1mm]
\vspace{1mm}

\begin{minipage}[t]{0.48\textwidth}\centering
  \includegraphics[width=\linewidth]{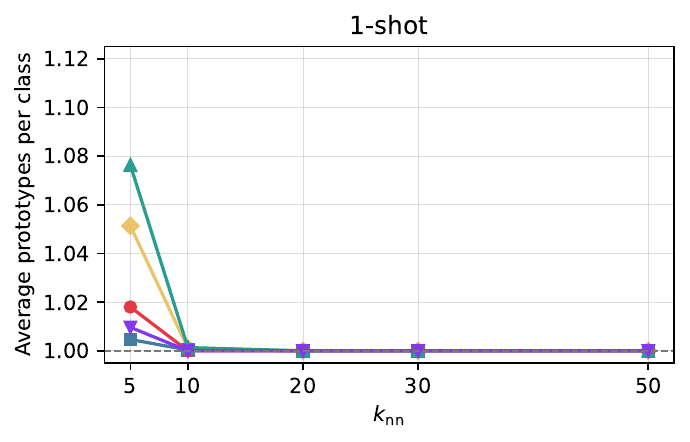}\\[-1mm]
  {\footnotesize (a) 5-way 1-shot.}
\end{minipage}\hfill
\begin{minipage}[t]{0.48\textwidth}\centering
  \includegraphics[width=\linewidth]{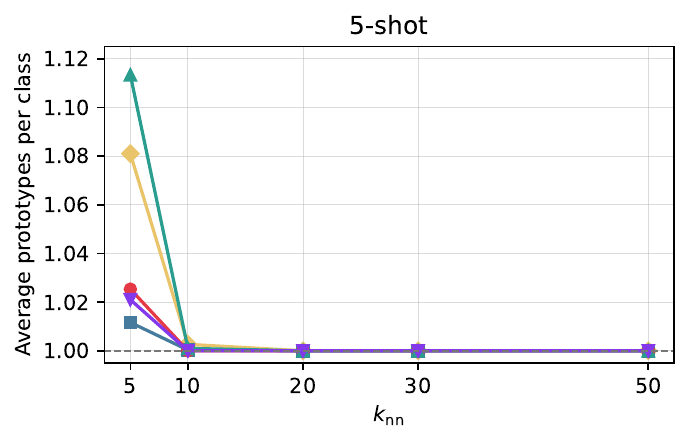}\\[-1mm]
  {\footnotesize (b) 5-way 5-shot.}
\end{minipage}
\caption{Average number of allocated prototypes per class as the graph neighbourhood size varies. Each curve corresponds to a dataset in the allocation sweep and averages over 600 episodes. Smaller neighbourhood graphs weakly activate additional prototypes, particularly for Omniglot and DTD, while the default $k_{\mathrm{nn}}=20$ and denser graphs reduce to one representative per class in all displayed datasets.}
\label{fig:knn_avg_prototypes}
\end{figure}

Figure~\ref{fig:knn_avg_prototypes} reveals a coupling between graph construction
and the multi-prototype extension that is not visible from the default setting alone.
At $k_{\mathrm{nn}}=5$, the average allocation rises above one prototype per class
on every displayed dataset, reaching $1.076$ prototypes per class for Omniglot
in 1-shot evaluation and $1.113$ in 5-shot evaluation. Averaged across the five
datasets, the corresponding values are $1.032$ and $1.050$. At
$k_{\mathrm{nn}}=10$ the activation is negligible, and for
$k_{\mathrm{nn}}\geq20$ all displayed allocations reduce exactly to one
representative per class.

This pattern indicates that prototype activation is sensitive to graph locality.
A sparse graph may break an episode into more weakly connected components and
thereby trigger additional capacity, even when those extra representatives are
not beneficial for recognition. Accordingly, this ablation does not revise our
main interpretation: the fixed STR gains are attributable to iterative refinement.
It does identify a concrete direction for further study: future experiments will
jointly analyse graph connectivity, verified intra-class multimodality, and
allocation calibration on episodes designed to require more than one
representative per class.

\subsection{Episode-adaptive graph connectivity}
\label{sec:dynamic_knn}

\begin{table}[H]
\centering
\papertablefont
\setlength{\tabcolsep}{7pt}
\renewcommand{\arraystretch}{1.07}
\caption{Episode-adaptive graph connectivity analysis. Dynamic-$k$NN selects a neighbourhood size for each episode; Fixed-$k$NN uses the shared configuration. Results are mean accuracy (\%) with 95\% confidence intervals over 600 episodes.}
\label{tab:dynamic_knn_delta}
\begin{tabular}{llccc}
\toprule
Dataset & Setting & Fixed-$k$NN & Dynamic-$k$NN & $\Delta$ \\
\midrule
miniImageNet & 1-shot & $87.55{\pm}0.69$ & $91.96{\pm}0.57$ & $\mathbf{+4.41}$ \\
 & 5-shot & $93.21{\pm}0.33$ & $95.79{\pm}0.24$ & $\mathbf{+2.58}$ \\
tieredImageNet & 1-shot & $92.58{\pm}0.62$ & $92.47{\pm}0.65$ & $-0.11$ \\
 & 5-shot & $96.31{\pm}0.34$ & $96.27{\pm}0.34$ & $-0.04$ \\
Caltech101 & 1-shot & $94.97{\pm}0.48$ & $92.86{\pm}0.52$ & $-2.11$ \\
 & 5-shot & $98.09{\pm}0.17$ & $97.17{\pm}0.25$ & $-0.92$ \\
FC100 & 1-shot & $58.08{\pm}1.10$ & $57.76{\pm}1.21$ & $-0.32$ \\
 & 5-shot & $68.30{\pm}0.88$ & $67.50{\pm}0.89$ & $-0.80$ \\
\bottomrule
\end{tabular}
\end{table}

We also evaluate an episode-adaptive graph strategy, Dynamic-$k$NN, which selects neighbourhood sizes per test episode. While Table~\ref{tab:dynamic_knn_delta} shows Dynamic-$k$NN substantially improves in-domain miniImageNet performance ($+4.41$ and $+2.58$ points for 1- and 5-shot), it does not generalize consistently. On tieredImageNet, the effect is negligible, and on Caltech-101 and FC100, adaptive construction reduces accuracy. This suggests that while adaptive graphs exploit local regularities in source distributions, they may over-adjust geometry under distribution shift. We thus retain a fixed $k_{\mathrm{nn}}$ for our main configuration.

\subsection{Scope of comparison and inference efficiency}
\label{sec:comparison_scope}

By shifting our primary cross-domain evaluation to a standard ResNet-10 backbone pretrained exclusively on the \textit{mini}ImageNet source domain, we align the feature-training protocol exactly with recent CD-FSL augmentation methods. The remaining differences between \method and these baselines are therefore confined to inference: \method performs no source-domain augmentation, no FFT or adversarial perturbation, and no encoder fine-tuning, and instead adapts only through transductive spectral refinement of the target episode.

We are explicit that this transductive setting differs from the inductive protocol of the cited methods. Consequently, the contribution we draw from Table~\ref{tab:crossdomain_8datasets} is a statement about \emph{where} robustness can be obtained---cheaply, at test time, from unlabeled target structure---rather than a fully setting-controlled ranking. Read this way, the results position spectral transductive inference as an efficient complement to, not a like-for-like substitute for, heavy source-domain regularization: when a frozen encoder already provides broadly separable target features, the joint episode geometry can be exploited at inference for substantial and, in the 1-shot regime, decisive gains.

\section{Limitations and Discussion}
\label{sec:limitations}
The experimental framing exposes limitations that matter for interpreting the results.
First, while \method operates as a highly efficient, training-free inference rule, it remains fundamentally bounded by the quality of the frozen representation. Because it does not fine-tune encoder weights, it cannot recover semantic features that were entirely discarded during source-domain pretraining; the ISIC result in Sec.~\ref{sec:cross_domain} is a concrete instance of this bound.
Second, STR assumes a closed-set transductive batch: all queries are available simultaneously and are expected to belong to the support classes. Realistic transductive evaluation may involve class imbalance or distractor queries~\cite{ren2018semisupervised,veilleux2021realistic}. Relatedly, our cross-domain comparison places a transductive method beside inductive baselines; we report this openly, and a fully setting-matched comparison against transductive CD-FSL methods is an important next step.
Third, spectral eigendecomposition introduces overhead compared with direct nearest-centroid inference; this is practical at standard episode sizes but must be reassessed as the number of ways or queries grows.
Fourth, pseudo-label refinement can amplify mistakes when the initial spectral partition is poor.
Finally, although the full pipeline includes a multi-prototype extension, it does not activate in the fixed reported configuration. The sparse-graph sweep indicates that activation can be induced by graph locality, but not that extra prototypes are beneficial; further study should therefore distinguish verified class multi-modality from graph fragmentation.

These limitations also clarify what the paper contributes. \method is not proposed as a replacement for end-to-end representation learning, but rather as a transparent, training-free mechanism for studying how unlabeled target geometry can refine prototypes under a frozen representation, accompanied by diagnostics that identify when a more elaborate claimed mechanism is inactive.

\section{Conclusion}
\label{sec:conclusion}
We presented \method, a training-free spectral transductive refinement procedure for few-shot classification with frozen visual features. \method embeds the complete support--query episode using a normalized-Laplacian representation and iteratively updates class representatives from pseudo-labelled queries.

Under a controlled ResNet-18 component study, the method shows its clearest benefit in cross-domain 1-shot evaluation, where representative estimation is most fragile. Furthermore, under the standard \textit{mini}ImageNet-pretrained ResNet-10 protocol across eight CD-FSL benchmarks, \method attains the highest 1-shot average and remains highly competitive at 5-shot against recent training-time augmentation methods. We report our transductive setting transparently: robust cross-domain adaptation can be obtained cheaply at inference through joint query geometry, complementing expensive source-domain regularization. A complementary analysis shows sparse graphs weakly activate extra prototypes, motivating future study into graph construction and multi-modal class structure.

\bibliography{references}

\begin{thebibliography}{44}
\providecommand{\natexlab}[1]{#1}
\providecommand{\url}[1]{\texttt{#1}}
\expandafter\ifx\csname urlstyle\endcsname\relax
  \providecommand{\doi}[1]{doi: #1}\else
  \providecommand{\doi}{doi: \begingroup \urlstyle{rm}\Url}\fi

\bibitem[Allen et~al.(2019)Allen, Shelhamer, Shin, and Tenenbaum]{allen2019imp}
Kelsey Allen, Evan Shelhamer, Hanul Shin, and Joshua~B. Tenenbaum.
\newblock Infinite mixture prototypes for few-shot learning.
\newblock In \emph{Int. Conf. Mach. Learn.}, pages 232--241, 2019.

\bibitem[Basu et~al.(2024)Basu, Hu, Massiceti, and Feizi]{basu2024strong}
Samyadeep Basu, Shell Hu, Daniela Massiceti, and Soheil Feizi.
\newblock Strong baselines for parameter-efficient few-shot fine-tuning.
\newblock In \emph{Proc. AAAI Conf. Artif. Intell.}, pages 11024--11031, 2024.

\bibitem[Belkin and Niyogi(2003)]{belkin2003laplacian}
Mikhail Belkin and Partha Niyogi.
\newblock Laplacian eigenmaps for dimensionality reduction and data
  representation.
\newblock \emph{Neural Computation}, 15\penalty0 (6):\penalty0 1373--1396,
  2003.

\bibitem[Boudiaf et~al.(2020)Boudiaf, Ziko, Rony, Dolz, Piantanida, and
  Ben~Ayed]{boudiaf2020tim}
Malik Boudiaf, Imtiaz~Masud Ziko, J{\'e}r{\^o}me Rony, Jos{\'e} Dolz, Pablo
  Piantanida, and Ismail Ben~Ayed.
\newblock Transductive information maximization for few-shot learning.
\newblock In \emph{Adv. Neural Inf. Process. Syst.}, 2020.

\bibitem[Chapelle et~al.(1999)Chapelle, Vapnik, and
  Weston]{chapelle1999transductive}
Olivier Chapelle, Vladimir Vapnik, and Jason Weston.
\newblock Transductive inference for estimating values of functions.
\newblock In \emph{Adv. Neural Inf. Process. Syst.}, pages 421--427, 1999.

\bibitem[Chen et~al.(2021)Chen, Liu, Xu, Darrell, and
  Wang]{chen2021metabaseline}
Yinbo Chen, Zhuang Liu, Huijuan Xu, Trevor Darrell, and Xiaolong Wang.
\newblock Meta-baseline: Exploring simple meta-learning for few-shot learning.
\newblock In \emph{Proc. IEEE/CVF Int. Conf. Comput. Vis.}, 2021.

\bibitem[Cimpoi et~al.(2014)Cimpoi, Maji, Kokkinos, Vedaldi, and
  Soatto]{cimpoi2014dtd}
Mircea Cimpoi, Subhransu Maji, Iasonas Kokkinos, Andrea Vedaldi, and Stefano
  Soatto.
\newblock Describing textures in the wild.
\newblock In \emph{Proc. IEEE/CVF Conf. Comput. Vis. Pattern Recognit.}, 2014.

\bibitem[Fei-Fei et~al.(2004)Fei-Fei, Fergus, and Perona]{feifei2004caltech}
Li~Fei-Fei, Rob Fergus, and Pietro Perona.
\newblock Learning generative visual models from few training examples: An
  incremental bayesian approach tested on 101 object categories.
\newblock In \emph{CVPR Workshop}, 2004.

\bibitem[Finn et~al.(2017)Finn, Abbeel, and Levine]{finn2017maml}
Chelsea Finn, Pieter Abbeel, and Sergey Levine.
\newblock Model-agnostic meta-learning for fast adaptation of deep networks.
\newblock In \emph{Int. Conf. Mach. Learn.}, 2017.

\bibitem[Fu et~al.(2023)Fu, Xie, Fu, and Jiang]{fu2023styleadv}
Yuqian Fu, Yu~Xie, Yanwei Fu, and Yu-Gang Jiang.
\newblock Styleadv: Meta style adversarial training for cross-domain few-shot
  learning.
\newblock In \emph{Proc. IEEE/CVF Conf. Comput. Vis. Pattern Recognit.}, 2023.

\bibitem[Garcia and Bruna(2018)]{garcia2018few}
Victor Garcia and Joan Bruna.
\newblock Few-shot learning with graph neural networks.
\newblock In \emph{Int. Conf. Learn. Represent.}, 2018.

\bibitem[He et~al.(2016)He, Zhang, Ren, and Sun]{he2016deep}
Kaiming He, Xiangyu Zhang, Shaoqing Ren, and Jian Sun.
\newblock Deep residual learning for image recognition.
\newblock In \emph{Proc. IEEE/CVF Conf. Comput. Vis. Pattern Recognit.}, pages
  770--778, 2016.

\bibitem[Helber et~al.(2019)Helber, Bischke, Dengel, and
  Borth]{helber2019eurosat}
Patrick Helber, Benjamin Bischke, Andreas Dengel, and Damian Borth.
\newblock Eurosat: A novel dataset and deep learning benchmark for land use and
  land cover classification.
\newblock \emph{IEEE J. Sel. Topics Appl. Earth Observ. Remote Sens.},
  12\penalty0 (7):\penalty0 2217--2226, 2019.

\bibitem[Hu et~al.(2022)Hu, Li, St\"uhmer, Kim, and Hospedales]{hu2022pmf}
Shell~Xu Hu, Da~Li, Jan St\"uhmer, Minyoung Kim, and Timothy~M. Hospedales.
\newblock Pushing the limits of simple pipelines for few-shot learning:
  External data and fine-tuning make a difference.
\newblock In \emph{Proc. IEEE/CVF Conf. Comput. Vis. Pattern Recognit.}, 2022.

\bibitem[Iscen et~al.(2019)Iscen, Tolias, Avrithis, and
  Chum]{iscen2019labelprop}
Ahmet Iscen, Giorgos Tolias, Yannis Avrithis, and Ondrej Chum.
\newblock Label propagation for deep semi-supervised learning.
\newblock In \emph{Proc. IEEE/CVF Conf. Comput. Vis. Pattern Recognit.}, 2019.

\bibitem[Krause et~al.(2013)Krause, Stark, Deng, and Fei-Fei]{krause20133d}
Jonathan Krause, Michael Stark, Jia Deng, and Li~Fei-Fei.
\newblock 3d object representations for fine-grained categorization.
\newblock In \emph{Proc. IEEE/CVF Int. Conf. Comput. Vis.}, 2013.

\bibitem[Lake et~al.(2015)Lake, Salakhutdinov, and Tenenbaum]{lake2015human}
Brenden~M. Lake, Ruslan Salakhutdinov, and Joshua~B. Tenenbaum.
\newblock Human-level concept learning through probabilistic program induction.
\newblock \emph{Science}, 350\penalty0 (6266):\penalty0 1332--1338, 2015.

\bibitem[Li et~al.(2025)Li, Fang, and Xue]{li2025hap}
Wenqian Li, Pengfei Fang, and Hui Xue.
\newblock Hap: Harmonized amplitude perturbation for cross-domain few-shot
  learning.
\newblock In \emph{Proc. AAAI Conf. Artif. Intell.}, 2025.

\bibitem[Li et~al.(2022)Li, Dai, Ge, Liu, Shan, and Duan]{li2022uncertainty}
Xiaomeng Li, Yongqiang Dai, Yixiao Ge, Jianzhuang Liu, Ying Shan, and Lingyu
  Duan.
\newblock Uncertainty modeling for out-of-distribution generalization.
\newblock In \emph{Int. Conf. Learn. Represent.}, 2022.

\bibitem[Liu et~al.(2019)Liu, Lee, Park, Kim, Yang, Hwang, and
  Yang]{liu2019tpn}
Yanbin Liu, Juho Lee, Minseop Park, Saehoon Kim, Eunho Yang, Sung~Ju Hwang, and
  Yi~Yang.
\newblock Learning to propagate labels: Transductive propagation network for
  few-shot learning.
\newblock In \emph{Int. Conf. Learn. Represent.}, 2019.

\bibitem[Mohanty et~al.(2016)Mohanty, Hughes, and
  Salath{\'e}]{mohanty2016using}
Sharada~P Mohanty, David~P Hughes, and Marcel Salath{\'e}.
\newblock Using deep learning for image-based plant disease detection.
\newblock \emph{Frontiers in plant science}, 7:\penalty0 1419, 2016.

\bibitem[Ng et~al.(2002)Ng, Jordan, and Weiss]{ng2002spectral}
Andrew~Y. Ng, Michael~I. Jordan, and Yair Weiss.
\newblock On spectral clustering: Analysis and an algorithm.
\newblock In \emph{Adv. Neural Inf. Process. Syst.}, 2002.

\bibitem[Nilsback and Zisserman(2008)]{nilsback2008flowers}
Maria-Elena Nilsback and Andrew Zisserman.
\newblock Automated flower classification over a large number of classes.
\newblock In \emph{Indian Conf. Comput. Vis., Graph. Image Process.}, pages
  722--729, 2008.

\bibitem[Oreshkin et~al.(2018)Oreshkin, Rodr{\'i}guez~L{\'o}pez, and
  Lacoste]{oreshkin2018tadam}
Boris~N. Oreshkin, Pau Rodr{\'i}guez~L{\'o}pez, and Alexandre Lacoste.
\newblock Tadam: Task dependent adaptive metric for improved few-shot learning.
\newblock In \emph{Adv. Neural Inf. Process. Syst.}, 2018.

\bibitem[Ren et~al.(2018)Ren, Triantafillou, Ravi, Snell, Swersky, Tenenbaum,
  Larochelle, and Zemel]{ren2018semisupervised}
Mengye Ren, Eleni Triantafillou, Sachin Ravi, Jake Snell, Kevin Swersky,
  Joshua~B. Tenenbaum, Hugo Larochelle, and Richard~S. Zemel.
\newblock Meta-learning for semi-supervised few-shot classification.
\newblock In \emph{Int. Conf. Learn. Represent.}, 2018.

\bibitem[Shen et~al.(2021)Shen, Xiao, Hu, Sbai, and Aubry]{shen2021reranking}
Xi~Shen, Yang Xiao, Shell~Xu Hu, Othman Sbai, and Mathieu Aubry.
\newblock Re-ranking for image retrieval and transductive few-shot
  classification.
\newblock In \emph{Adv. Neural Inf. Process. Syst.}, 2021.

\bibitem[Shi and Malik(2000)]{shi2000normalized}
Jianbo Shi and Jitendra Malik.
\newblock Normalized cuts and image segmentation.
\newblock \emph{IEEE Trans. Pattern Anal. Mach. Intell.}, 22\penalty0
  (8):\penalty0 888--905, 2000.

\bibitem[Snell et~al.(2017)Snell, Swersky, and Zemel]{snell2017prototypical}
Jake Snell, Kevin Swersky, and Richard~S. Zemel.
\newblock Prototypical networks for few-shot learning.
\newblock In \emph{Adv. Neural Inf. Process. Syst.}, 2017.

\bibitem[Tian et~al.(2020)Tian, Wang, Krishnan, Tenenbaum, and
  Isola]{tian2020rethinking}
Yonglong Tian, Yue Wang, Dilip Krishnan, Joshua~B. Tenenbaum, and Phillip
  Isola.
\newblock Rethinking few-shot image classification: A good embedding is all you
  need?
\newblock In \emph{European Conf. Comput. Vis.}, 2020.

\bibitem[Tschandl et~al.(2018)Tschandl, Rosendahl, and
  Kittler]{tschandl2018ham10000}
Philipp Tschandl, Cliff Rosendahl, and Harald Kittler.
\newblock The ham10000 dataset, a large collection of multi-source
  dermatoscopic images of common pigmented skin lesions.
\newblock \emph{Scientific data}, 5\penalty0 (1):\penalty0 1--9, 2018.

\bibitem[Van~Horn et~al.(2018)Van~Horn, Mac~Aodha, Song, Cui, Sun, Shepard,
  Adam, Perona, and Belongie]{van2018inaturalist}
Grant Van~Horn, Oisin Mac~Aodha, Yang Song, Yin Cui, Chen Sun, Alex Shepard,
  Hartmut Adam, Pietro Perona, and Serge Belongie.
\newblock The inaturalist species classification and detection dataset.
\newblock In \emph{Proc. IEEE/CVF Conf. Comput. Vis. Pattern Recognit.}, 2018.

\bibitem[Veilleux et~al.(2021)Veilleux, Boudiaf, Piantanida, and
  Ben~Ayed]{veilleux2021realistic}
Olivier Veilleux, Malik Boudiaf, Pablo Piantanida, and Ismail Ben~Ayed.
\newblock Realistic evaluation of transductive few-shot learning.
\newblock In \emph{Adv. Neural Inf. Process. Syst.}, 2021.

\bibitem[Vinyals et~al.(2016)Vinyals, Blundell, Lillicrap, Kavukcuoglu, and
  Wierstra]{vinyals2016matching}
Oriol Vinyals, Charles Blundell, Timothy Lillicrap, Koray Kavukcuoglu, and Daan
  Wierstra.
\newblock Matching networks for one shot learning.
\newblock In \emph{Adv. Neural Inf. Process. Syst.}, 2016.

\bibitem[von Luxburg(2007)]{vonluxburg2007tutorial}
Ulrike von Luxburg.
\newblock A tutorial on spectral clustering.
\newblock \emph{Statistics and Computing}, 17\penalty0 (4):\penalty0 395--416,
  2007.

\bibitem[Wah et~al.(2011)Wah, Branson, Welinder, Perona, and
  Belongie]{wah2011cub}
Catherine Wah, Steve Branson, Peter Welinder, Pietro Perona, and Serge
  Belongie.
\newblock The caltech-ucsd birds-200-2011 dataset.
\newblock Technical report, California Institute of Technology, 2011.

\bibitem[Wang and Deng(2021)]{wang2021cross}
Haoqing Wang and Zhi-Hong Deng.
\newblock Cross-domain few-shot classification via adversarial task
  augmentation.
\newblock In \emph{Proc. Int. Joint Conf. Artif. Intell. (IJCAI)}, 2021.

\bibitem[Wang et~al.(2017)Wang, Peng, Lu, Lu, Bagheri, and
  Summers]{wang2017chestx}
Xiaosong Wang, Yifan Peng, Le~Lu, Zhiyong Lu, Mohammadhadi Bagheri, and
  Ronald~M Summers.
\newblock Chestx-ray8: Hospital-scale chest x-ray database and benchmarks on
  weakly-supervised classification and localization of common thorax diseases.
\newblock In \emph{Proc. IEEE/CVF Conf. Comput. Vis. Pattern Recognit.}, 2017.

\bibitem[Wang et~al.(2019)Wang, Chao, Weinberger, and van~der
  Maaten]{wang2019simpleshot}
Yan Wang, Wei-Lun Chao, Kilian~Q. Weinberger, and Laurens van~der Maaten.
\newblock Simpleshot: Revisiting nearest-neighbor classification for few-shot
  learning.
\newblock \emph{arXiv preprint arXiv:1911.04623}, 2019.

\bibitem[Zhou et~al.(2017)Zhou, Lapedriza, Khosla, Oliva, and
  Torralba]{zhou2017places}
Bolei Zhou, Agata Lapedriza, Aditya Khosla, Aude Oliva, and Antonio Torralba.
\newblock Places: A 10 million image database for scene recognition.
\newblock \emph{IEEE Trans. Pattern Anal. Mach. Intell.}, 40\penalty0
  (6):\penalty0 1452--1464, 2017.

\bibitem[Zhou et~al.(2004)Zhou, Bousquet, Lal, Weston, and
  Sch{\"o}lkopf]{zhou2004learning}
Dengyong Zhou, Olivier Bousquet, Thomas~N. Lal, Jason Weston, and Bernhard
  Sch{\"o}lkopf.
\newblock Learning with local and global consistency.
\newblock In \emph{Adv. Neural Inf. Process. Syst.}, 2004.

\bibitem[Zhou et~al.(2024)Zhou, Wang, Zhang, Ding, Lin, Chen, Wei, and
  Zhang]{zhou2024meta}
Fei Zhou, Peng Wang, Lei Zhang, Chen Ding, Guosheng Lin, Zhenghua Chen, Wei
  Wei, and Yanning Zhang.
\newblock Meta-exploiting frequency prior for cross-domain few-shot learning.
\newblock In \emph{Adv. Neural Inf. Process. Syst.}, 2024.

\bibitem[Zhu et~al.(2003)Zhu, Ghahramani, and Lafferty]{zhu2003semi}
Xiaojin Zhu, Zoubin Ghahramani, and John~D. Lafferty.
\newblock Semi-supervised learning using gaussian fields and harmonic
  functions.
\newblock In \emph{Int. Conf. Mach. Learn.}, 2003.

\bibitem[Ziko et~al.(2020)Ziko, Dolz, Granger, and Ben~Ayed]{ziko2020laplacian}
Imtiaz~Masud Ziko, Jos{\'e} Dolz, Eric Granger, and Ismail Ben~Ayed.
\newblock Laplacian regularized few-shot learning.
\newblock In \emph{Int. Conf. Mach. Learn.}, pages 11660--11670, 2020.

\bibitem[Zou et~al.(2024)Zou, Liu, Hu, Li, and Li]{zou2024flatten}
Yiru Zou, Yuhang Liu, Yulei Hu, Ying Li, and Rui Li.
\newblock Flatten long-range loss landscapes for cross-domain few-shot
  learning.
\newblock In \emph{Proc. IEEE/CVF Conf. Comput. Vis. Pattern Recognit.}, 2024.

\end{thebibliography}
\end{document}


\maketitle

\section{Additional Domain-Shift Check}
\label{sec:supp_additional_transfer}
\begin{table}[t]
\centering
\papertablefont
\setlength{\tabcolsep}{4pt}
\renewcommand{\arraystretch}{1.07}
\caption{Additional domain-shift check using frozen ImageNet-1K pretrained
ResNet-18 features. Results are mean accuracy (\%) with 95\% confidence
intervals over 600 episodes.}
\label{tab:additional_transfer}
\begin{tabular}{lrrr}
\toprule
Dataset & Shot & Spectral init. & STR \\
\midrule
EuroSAT & 1 & 72.32$\pm$0.92 & 76.32$\pm$1.06 \\
 & 5 & 84.82$\pm$0.59 & 85.63$\pm$0.61 \\
\bottomrule
\end{tabular}
\end{table}

EuroSAT extends the evaluation to satellite imagery, a visually distinct
target domain relative to standard natural-image recognition. STR improves over
spectral initialization in both shot settings under the same frozen ResNet-18
representation protocol. This result complements the cross-domain component
study in the main paper and is interpreted as within-framework evidence rather
than as a cross-method ranking.

\section{Transfer to Frozen DINOv2 Representations}
\label{sec:supp_dinov2}
The inference procedure is also compatible with stronger frozen visual
representations. We therefore evaluate it with DINOv2 ViT-S/14
embeddings~\cite{oquab2023dinov2}. DINOv2 uses large-scale external
self-supervised pretraining and is consequently kept separate from the
backbone-grouped standard-benchmark context in the main paper.

\begin{table}[t]
\centering
\papertablefont
\setlength{\tabcolsep}{3.3pt}
\renewcommand{\arraystretch}{1.07}
\caption{Transfer evaluation using frozen DINOv2 ViT-S/14 features. Results
are mean accuracy (\%) with 95\% confidence intervals over 600 episodes.
This study evaluates compatibility with foundation features; recorded graph
settings are shown explicitly.}
\label{tab:dinov2_transfer}
\begin{tabular}{llccc}
\toprule
Dataset & Variant & $k_{\mathrm{nn}}$ & 1-shot & 5-shot \\
\midrule
miniImageNet & Spectral initialization & 20 & $91.04{\pm}0.46$ & $97.01{\pm}0.25$ \\
 & Full refinement & 20 & $\mathbf{92.43{\pm}0.48}$ & $\mathbf{97.56{\pm}0.24}$ \\
tieredImageNet & Spectral initialization & 10 & $88.88{\pm}0.61$ & $95.45{\pm}0.37$ \\
 & Full refinement & 20 & $\mathbf{90.85{\pm}0.54}$ & $\mathbf{96.75{\pm}0.35}$ \\
\bottomrule
\end{tabular}
\end{table}

The miniImageNet results use a matched graph configuration and show that
full refinement improves spectral initialization under DINOv2 features.
For tieredImageNet, the initialization and refinement rows use
different neighbourhood settings, stated explicitly in the table; they are
therefore evidence of encoder compatibility rather than a controlled component
isolation.

\section{Five-Shot Sensitivity Analysis}
\label{sec:supp_sensitivity}
\begin{figure*}[t]
\centering
\includegraphics[width=0.68\textwidth]{figures/ablations/shared_legend.pdf}\\[-1mm]
\vspace{1mm}
\begin{minipage}[t]{0.32\textwidth}\centering
 \includegraphics[width=\linewidth]{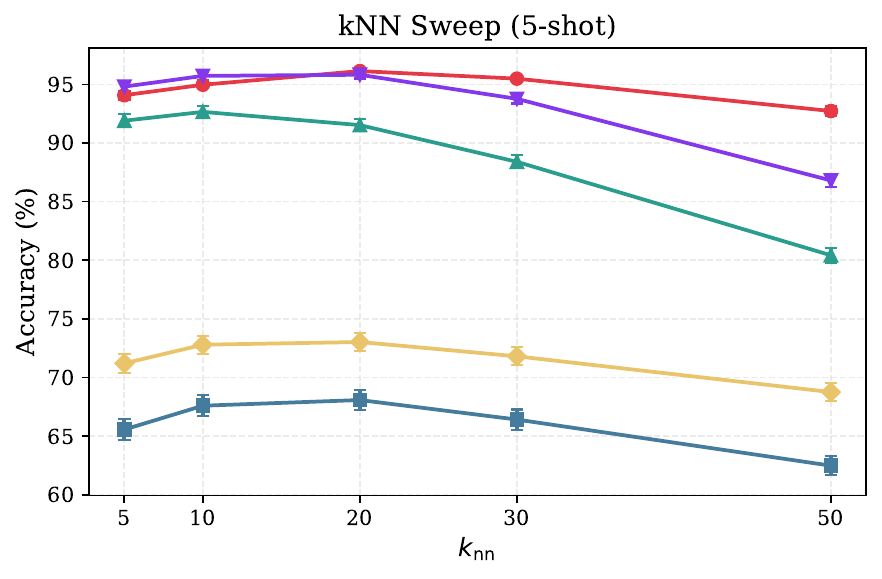}\\[-1mm]
 {\footnotesize (a) $k_{\mathrm{nn}}$ sweep.}
\end{minipage}\hfill
\begin{minipage}[t]{0.32\textwidth}\centering
 \includegraphics[width=\linewidth]{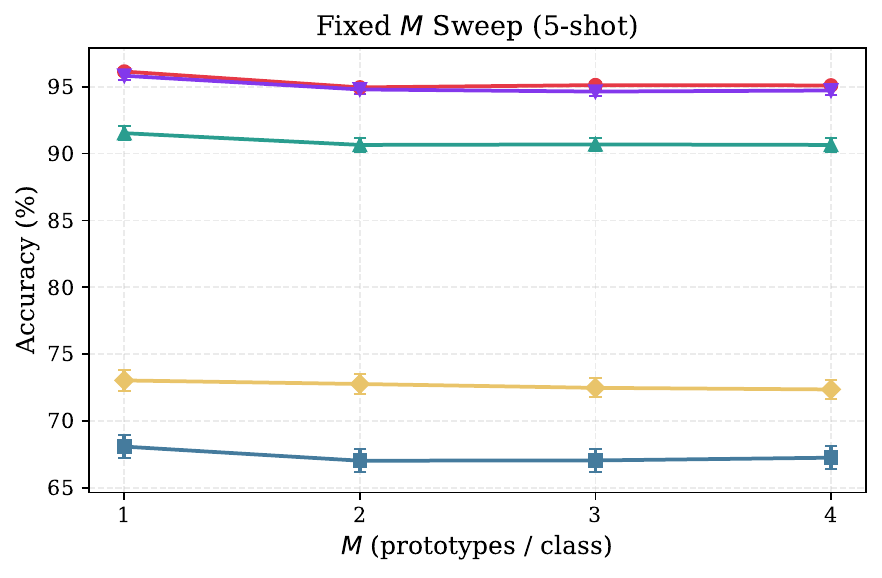}\\[-1mm]
 {\footnotesize (b) Fixed prototype count.}
\end{minipage}\hfill
\begin{minipage}[t]{0.32\textwidth}\centering
 \includegraphics[width=\linewidth]{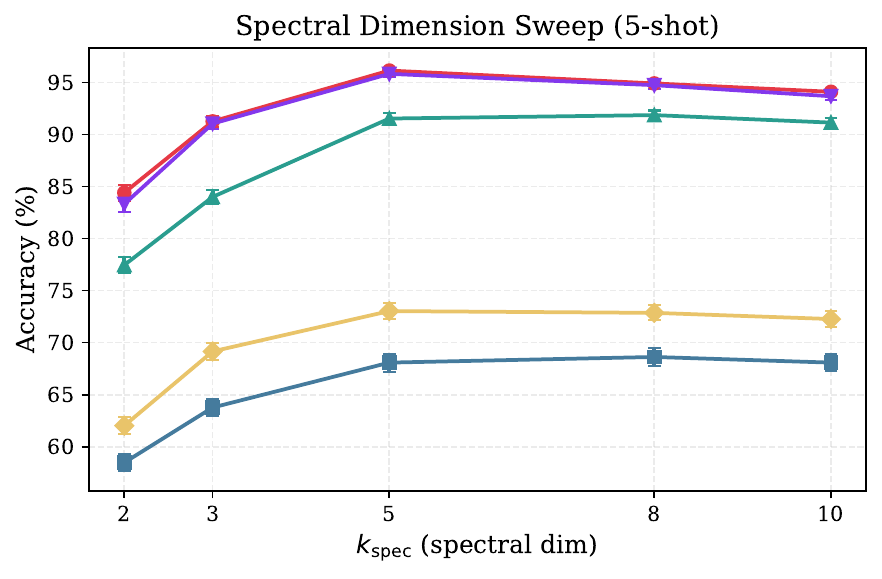}\\[-1mm]
 {\footnotesize (c) Spectral dimension.}
\end{minipage}
\caption{Five-shot sensitivity analysis with frozen ResNet-18 features over the
component-study datasets. Compared with the one-shot results in the main paper,
the five-shot regime is less sensitive because multiple labelled supports already
provide a more stable initial estimate of each class representative.}
\label{fig:supp_sweeps_five}
\end{figure*}

The main manuscript reports one-shot sensitivity analysis, where STR gains are
most pronounced. For completeness, Fig.~\ref{fig:supp_sweeps_five} presents
the corresponding five-shot analysis under frozen ResNet-18 features.

The five-shot plots reinforce that moderate graph locality and a compact
spectral representation remain preferable to extreme constructions. Additional
fixed prototypes do not yield consistent improvements, supporting the
interpretation that the main benefit arises from query-guided refinement rather
than increased prototype capacity.

\section{Episode-Level Refinement Diagnostics}
\label{sec:supp_diagnostics}
\begin{figure}[t]
\centering
\includegraphics[width=.86\linewidth,height=.37\textheight,keepaspectratio]
{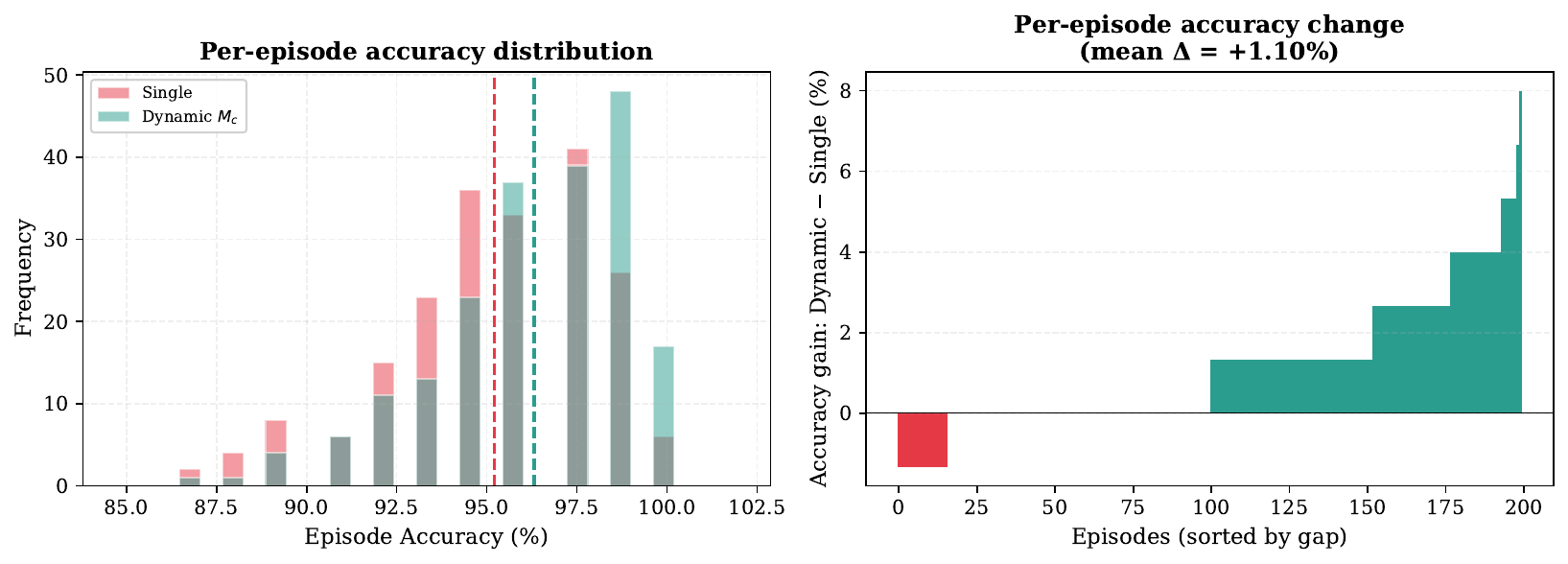}
\caption{Episode-level accuracy distributions before and after spectral
transductive refinement. The refined configuration improves the distribution
of episode accuracies in the analysed regime. The allocation diagnostic in
Fig.~\ref{fig:prototype_heatmap} shows that these episodes operate in the
single-representative STR regime.}
\label{fig:accuracy_distribution}
\end{figure}

\begin{figure}[p]
\centering
\vspace*{\fill}
\includegraphics[width=\linewidth,height=0.82\textheight,keepaspectratio]
{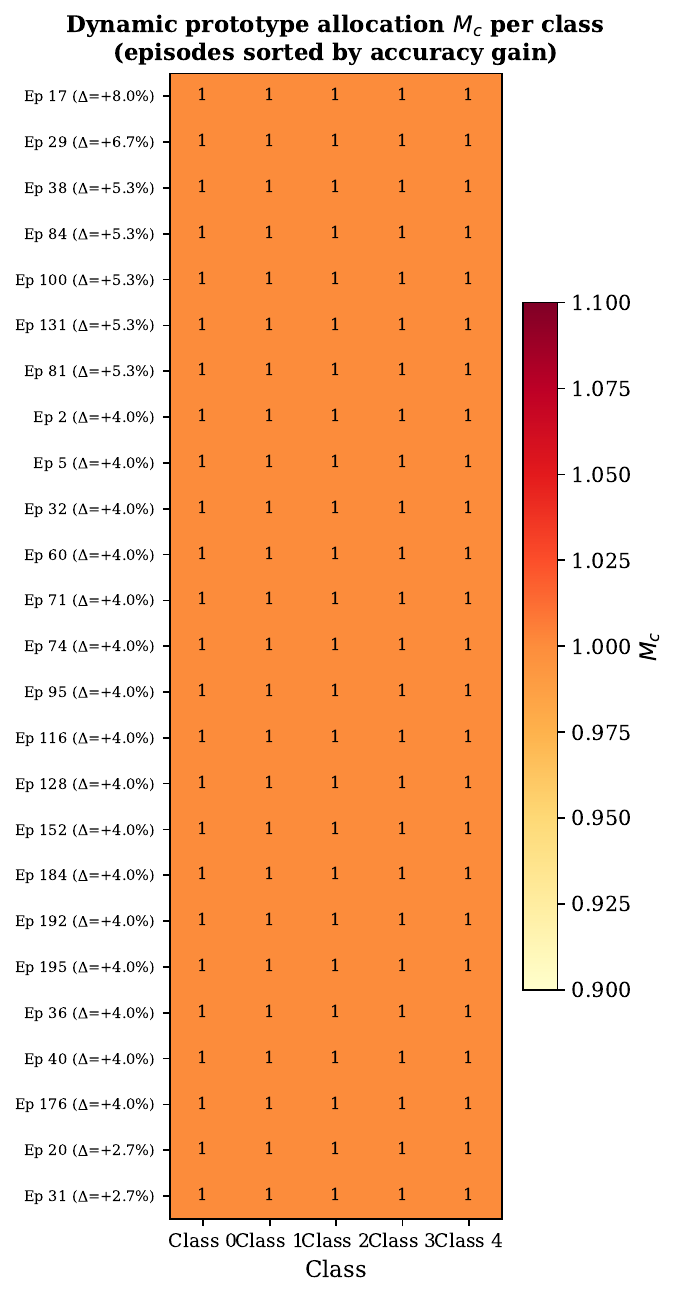}
\caption{Episode-level prototype-allocation diagnostic. Each cell reports the
number of representatives assigned to a class in an analysed episode. The
uniform value of one across the displayed episodes shows that, in this standard
evaluation regime, the adaptive-allocation extension reduces to iterative
single-representative spectral refinement. This supports attributing measured
improvements to STR rather than to additional prototype capacity.}
\label{fig:prototype_heatmap}
\vspace*{\fill}
\end{figure}

Figure~\ref{fig:accuracy_distribution} visualizes the distribution of episode
accuracies before and after full spectral refinement. The prototype-count
diagnostic in Fig.~\ref{fig:prototype_heatmap} shows that the displayed standard
episodes use one representative per class; the observed changes should therefore
be interpreted as iterative spectral representative refinement.

The prototype-allocation heatmap is placed at the end of this supplementary
material as a full-page diagnostic to preserve readability of its tall
class-by-episode layout.

\bibliography{references}

\clearpage
\clearpage